\documentclass{article}
\usepackage[margin=1in]{geometry}
\usepackage[numbers]{natbib}
\usepackage{url}

\usepackage[utf8]{inputenc}
\usepackage[T1]{fontenc}
\usepackage{amsmath,amssymb,amsfonts,amsthm}
\usepackage{graphicx}
\usepackage{booktabs}
\usepackage{multirow}
\usepackage{array}
\usepackage{tabularx}
\usepackage{caption}
\usepackage{subcaption}
\usepackage{xcolor}
\usepackage{enumitem}
\usepackage{hyperref}
\hypersetup{hidelinks}

\newtheorem{definition}{Definition}
\newtheorem{theorem}{Theorem}

\newcommand{\Rhat}{\widehat{R}}

\newcommand{\legenddash}{\rule[0.55ex]{0.32em}{0.4pt}\hspace{0.16em}\rule[0.55ex]{0.32em}{0.4pt}\hspace{0.16em}\rule[0.55ex]{0.32em}{0.4pt}}
\newcolumntype{Y}{>{\raggedright\arraybackslash}X}

\title{PUe: Biased Positive-Unlabeled Learning Enhancement by Causal Inference}
\date{}

\author{
Xutao Wang \quad Hanting Chen \quad Tianyu Guo \quad Yunhe Wang\thanks{Corresponding author.}\\
Huawei Noah's Ark Lab\\
\texttt{\{xutao.wang,chenhanting,tianyu.guo,yunhe.wang\}@huawei.com}
}

\begin{document}
\maketitle

\begin{abstract}
Positive-Unlabeled (PU) learning aims to achieve high-accuracy binary classification with limited labeled positive examples and numerous unlabeled ones. Existing cost-sensitive-based methods often rely on strong assumptions that examples with an observed positive label were selected entirely at random. In fact, the uneven distribution of labels is prevalent in real-world PU problems, indicating that most actual positive and unlabeled data are subject to selection bias. Building on the SAR-PU propensity-weighted framework of Bekker et al.~\cite{bekker2019beyond}, we study a PU learning enhancement (PUe) framework using normalized propensity scores and normalized inverse probability weighting (NIPW). PUe's main contributions are a normalized inverse-probability-weighted PU risk formulation; additional theoretical analyses of normalized sample-weight error and common PU estimators under biased labeling; regularized deep propensity-score estimation; integration with modern cost-sensitive PU methods; and support for selectively labeled negative classes. Experiments on MNIST, CIFAR-10, and ADNI demonstrate improvements over several PU baselines under non-uniform label distributions. Codes are available at \href{https://github.com/huawei-noah/Noah-research/tree/master/PUe}{GitHub} and \href{https://gitee.com/mindspore/models/tree/master/research/cv/PUe}{Gitee}.
\end{abstract}

\section{Introduction}

In the era of big data, deep neural networks have achieved outstanding performance across various tasks, even surpassing human performance in many instances, particularly in traditional binary classification problems. The success of these deep neural networks often hinges on supervised learning using large quantities of labeled data. However, in reality, acquiring even binary labels can be challenging. For instance, in recommendation systems, users' multiple clicks on films may be considered as positive samples. Nonetheless, all other films cannot be assumed uninteresting and thus should not be treated as negative examples; instead, they should be regarded as unlabeled ones.

The same issue emerges in text classification, where it is typically more straightforward to define a partial set of positive samples. However, due to the vast diversity of negative samples, it becomes difficult or even impossible to describe a comprehensive set of negative samples that represent all content not included in the positive samples. Similar situations occur in medical diagnostics, malicious URL detection, and spam detection, where only a few labeled positives are available amidst a plethora of unlabeled data. This scenario is a variant of the classical binary classification setup, known as PU. In recent years, there has been a growing interest in this setting. Positive-Unlabeled (PU) learning primarily addresses the challenge of learning binary classifiers solely from positive samples and unlabeled data.

In previous research, numerous PU algorithms have been developed, with cost-sensitive PU learning emerging as a popular research direction. Methods such as \cite{liu2002partially,liu2003building,elkan2008learning} reweight positive and negative risks by hyper-parameters and minimize it. In addition, Self-PU~\cite{chen2020selfpu} introduces self-supervision to nnPU through auxiliary tasks, including model calibration and distillation; ImbPU~\cite{su2021positive} oversamples and modifies sample weights to address unbalanced data. Dist-PU~\cite{zhao2022distpu} corrects the negative preference of the classification model through prior information. PUSB~\cite{kato2019learning} maintains the order-preserving assumption. However, these methods necessitate the assumption that their checked set is uniformly sampled from the population, or else the PU learning risk estimator ceases to be an unbiased or consistent estimator, otherwise resulting in reduced model accuracy.

In reality, the labeled set is often biased and does not conform to the Selected Completely At Random (SCAR) assumption~\cite{hyttinen2013experiment,bekker2019beyond}, which posits that the observed labeled examples are a random subset of the complete set of positive examples. Consequently, it is essential to relax the assumption of the labeled set and replace it with a more general assumption about the labeling mechanism: the probability of selecting positive examples to be labeled depends on their attribute values, known as the Selected At Random (SAR) assumption~\cite{hyttinen2013experiment}.

The most directly related prior work is Bekker et al.~\cite{bekker2019beyond}, which studies positive-unlabeled learning beyond SCAR under the Selected At Random (SAR) labeling mechanism. PUe adapts this SAR-PU foundation to a normalized PU risk formulation, develops additional theoretical analyses for normalized weighting and common PU estimators under biased labeling, estimates propensity scores with regularized deep models, and integrates the resulting correction with modern cost-sensitive PU methods. A detailed component-level comparison between Bekker et al.~\cite{bekker2019beyond} and PUe is provided in Appendix~D.

Gerych et al.~\cite{gerych2022recovering} is also related, with a different emphasis on conditions for recovering identifiable propensity scores. That work discusses Local Certainty and Probabilistic Gap assumptions. Under the Local Certainty hypothesis, there is no overlap between the positive and negative hypotheses, $p(x^{\widehat{P}}\mid y=0)=0$. Under the Probabilistic Gap assumption, it is assumed that $e=k p(y=1\mid x)$ is linear and there is an anchor point. We use this work primarily to motivate the challenges of propensity-score estimation.

\begin{figure}[t]
\centering
\small
\begin{tabular}{@{}ll@{\qquad}ll@{}}
\textcolor[RGB]{111,205,220}{\large$\bullet$} & Labeled Positive Samples
& {\large$\circ$} & Unlabeled Negative Samples\\
\textcolor{gray}{\large$\bullet$} & Unlabeled Positive Samples
& \legenddash & Classification Boundary
\end{tabular}
\vspace{0.4em}

\begin{subfigure}{0.36\linewidth}
  \centering
  \includegraphics[width=\linewidth]{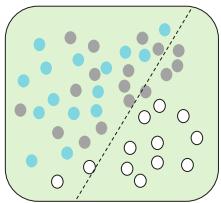}
  \caption{Traditional PU}
\end{subfigure}
\hspace{0.08\linewidth}
\begin{subfigure}{0.36\linewidth}
  \centering
  \includegraphics[width=\linewidth]{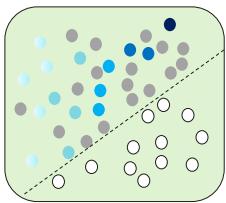}
  \caption{Proposed PUe}
\end{subfigure}
\caption{Our method and traditional PU method classification diagram. PUe uses reweighting to make the classification plane more accurate.}
\label{fig:classification}
\end{figure}

However, the above assumptions are too strict. We show in the appendix that the estimated propensity scores cannot be completely unbiased. To address the aforementioned problem, we propose a causal-inference-inspired PU learning framework, termed PUe. Our approach uses propensity weighting to correct the original PU learning risk estimator under biased conditions; PUe formulates this correction using normalized inverse probability weights. Given that propensity scores for samples are typically unknown, we apply regularization techniques to a deep learning classifier to estimate the propensity score of each sample in the labeled set. An illustration of the proposed method is shown in Figure~\ref{fig:classification}. PUe further analyzes normalized sample-weight error and common PU estimators under biased labeling, and integrates normalized propensity correction with modern cost-sensitive PU algorithms.

Our main contributions are summarized as follows:
\begin{itemize}[leftmargin=*]
  \item We formulate PUe with normalized inverse probability weighting in the PU risk formulation for biased positive-unlabeled learning.
  \item We develop additional theoretical analyses of normalized sample-weight error and common PU estimators under biased labeling.
  \item We extend propensity-score estimation to deep neural networks and introduce regularization to reduce overfitting and degenerate propensity estimates.
  \item We integrate the normalized propensity-weighting mechanism with cost-sensitive PU methods, including uPU, nnPU, PUbN, and Dist-PU.
  \item We extend the framework to selectively labeled negative classes, resulting in PUbNe.
  \item We evaluate the resulting framework on MNIST, CIFAR-10, and the Alzheimer's Disease Neuroimaging Initiative (ADNI) database.
\end{itemize}

\section{Methodology}

In this section, we review existing PU algorithms and their limitations in biased labeling scenarios, then introduce PUe to improve PU learning under biased labeling.

\subsection{Review of PU classification}

In standard PN classification, let $x\in\mathbb{R}^{d}$ and $y\in\{+1,-1\}$ be the input sample and its corresponding label. We are given positive data and negative data sampled independently from $p_P(x)=p(x\mid y=+1)$ and $p_N(x)=p(x\mid y=-1)$ as $\chi_P=\{x_i^P\}_{i=1}^{n_P}$ and $\chi_N=\{x_i^N\}_{i=1}^{n_N}$. We denote the class prior probability as $\pi=p(y=1)$ and follow the convention that $\pi$ is known throughout the paper~\cite{kiryo2017positive}. Class prior can also be estimated under biased positives and unlabeled examples~\cite{jain2020class}. Let $g:\mathbb{R}^{d}\to\mathbb{R}$ be the binary classifier, $\theta$ be its parameter, and $L:\mathbb{R}\times\{+1,-1\}\to\mathbb{R}_{+}$ be a loss function.

Let $R_P(g,+1)=\mathbb{E}_{x\sim p_P(x)}[L(g(x),+1)]$ and $R_N(g,-1)=\mathbb{E}_{x\sim p_N(x)}[L(g(x),-1)]$. The empirical PN risk is
\begin{equation}
\Rhat_{PN}(g)=\pi\Rhat_P(g,+1)+(1-\pi)\Rhat_N(g,-1),
\label{eq:pn-risk}
\end{equation}
where
\[
\Rhat_P(g,+1)=\frac{1}{n_P}\sum_{i=1}^{n_P}L(g(x_i^P),+1),\quad
\Rhat_N(g,-1)=\frac{1}{n_N}\sum_{i=1}^{n_N}L(g(x_i^N),-1).
\]

In standard PU classification, instead of negative data $\chi_N$, we have unlabeled samples $\chi_U=\{x_i^U\}_{i=1}^{n_U}\sim p(x)$. Since $p(x)=\pi p_P(x)+(1-\pi)p_N(x)$, we have
\begin{equation}
(1-\pi)R_N(g,-1)=R_U(g,-1)-\pi R_P(g,-1).
\label{eq:pu-identity}
\end{equation}
The unbiased PU empirical risk is
\begin{equation}
\Rhat_{uPU}(g)=\pi\Rhat_P(g,+1)+\Rhat_U(g,-1)-\pi\Rhat_P(g,-1),
\label{eq:upu-risk}
\end{equation}
where
\[
\Rhat_P(g,-1)=\frac{1}{n_P}\sum_{i=1}^{n_P}L(g(x_i^P),-1),\quad
\Rhat_U(g,-1)=\frac{1}{n_U}\sum_{i=1}^{n_U}L(g(x_i^U),-1).
\]

In theory, the risk $(1-\pi)R_N(g,-1)=R_U(g,-1)-\pi R_P(g,-1)$ is non-negative. However, if $g$ is too flexible, $\Rhat_{PU}(\hat{g}_{PU})$ can become negative and the model can severely overfit the training data. The non-negative risk estimator for PU learning alleviates overfitting~\cite{kiryo2017positive}.

\subsection{PU classification in biased scenarios}

PU learning mostly assumes that all labeled samples are selected completely at random from all positive samples, which is called Selected Completely At Random (SCAR).

\begin{definition}[SCAR, selected completely at random~\cite{elkan2008learning}]
Labeled examples are selected completely at random, independent from their attributes, from the positive distribution. The propensity score $e(x)$, which is the probability of selecting a positive example, is constant and equal to the label frequency:
\begin{equation}
e(x)=p(s=1\mid x,y=1)=p(s=1\mid y=1)=c.
\label{eq:scar}
\end{equation}
\end{definition}

However, many PU learning applications suffer from labeling bias. The SCAR assumption does not conform to reality. For example, whether someone clicks on a sponsored search ad is influenced by the position in which it is placed. Similarly, whether a patient with a disease will see a doctor depends on socio-economic status and symptom severity. The Selected At Random (SAR) assumption is a more general assumption about the labeling mechanism: the probability of selecting positive examples to be labeled depends on attribute values~\cite{bekker2018positive}.

\begin{definition}[SAR, selected at random~\cite{bekker2019beyond}]
The labeling mechanism depends on the values of the attributes of the example. However, given the attribute values, it does not depend further on the probability of the example being positive. Instead of assuming a constant probability for all positive examples to be labeled, SAR assumes that the probability is a function of a subset of the example's attributes:
\[
e(x)=p(s=1\mid x,y=1).
\]
\end{definition}

Motivated by causal-inference methods for correcting selection bias, \cite{bekker2019beyond} introduced using propensity scores and inverse-probability weighting (IPW) to construct a normalized correction for biased PU learning~\cite{rosenbaum1983central,hirano2003efficient}. A crucial difference from the propensity score in standard causal inference is that the PU propensity score is conditioned on the class being positive. Since negative examples have zero probability of being labeled as positive, IPW is applied to the labeled positive component. For each labeled example $(x_i,s_i=1)$ with propensity score $e_i$, there are expected to be $1/e_i$ positive examples. We propose to enhance this framework by using normalized inverse propensity weighting (NIPW).

The positive risk can be decomposed by propensity level:
\begin{equation}
R_P(g,+1)
=\mathbb{E}_{x\sim p_P(x)}[L(g(x),+1)]
=\mathbb{E}_{c\in(0,1]}\left\{\mathbb{E}_{x\sim p_P(x)}[L(g(x),+1)\mid e(x)=c]\right\}.
\label{eq:pos-risk-decomp}
\end{equation}
The PUe positive-risk estimator with normalized inverse-propensity weights is
\begin{equation}
\Rhat_P^{e}(g,+1)=\sum_{i=1}^{n_P}\omega_i^P L(g(x_i^P),+1).
\label{eq:pue-pos-plus}
\end{equation}
Similarly,
\begin{equation}
\Rhat_P^{e}(g,-1)=\sum_{i=1}^{n_P}\omega_i^P L(g(x_i^P),-1).
\label{eq:pue-pos-minus}
\end{equation}
Here $e(x_i^P)$ is the true propensity score and $\widehat{e}(x_i^P)$ is its estimate. The normalized inverse weights are
\[
\omega_i^P
=\frac{1/e(x_i^P)}{\sum_{j=1}^{n_P}1/e(x_j^P)},\qquad
\widehat{\omega}_i^P
=\frac{1/\widehat{e}(x_i^P)}{\sum_{j=1}^{n_P}1/\widehat{e}(x_j^P)}.
\]
Thus
\[
\sum_{i=1}^{n_P}\omega_i^P=1,\qquad
\sum_{i=1}^{n_P}\widehat{\omega}_i^P=1.
\]

Existing propensity-score estimation methods such as SAR-EM in Bekker et al.~\cite{bekker2019beyond} can underestimate propensity scores of positive samples, producing co-directional bias or degenerate estimates in which labeled samples have propensity score close to 1 and unlabeled samples have propensity score close to 0. PUe uses normalization to reduce co-directional deviation and regularization to alleviate overfitting. To calculate the corrected PU loss in biased scenarios, we consider two cases: known true propensity scores and estimated propensity scores.

\paragraph{Case 1: known true propensity scores.}
The true PN risk with real class labels is
\begin{equation}
\begin{aligned}
R_{PN}(g\mid y)
&=\pi\Rhat_P(g,+1)+(1-\pi)\Rhat_N(g,-1)\\
&=\frac{1}{n}\sum_{i=1}^{n}
\left[y_iL(g(x_i),+1)+(1-y_i)L(g(x_i),-1)\right].
\end{aligned}
\label{eq:true-pn-risk}
\end{equation}

The following estimator adapts the propensity-weighted SAR-PU estimator of Bekker et al.~\cite{bekker2019beyond} using normalized weights and PUe risk notation.

\begin{definition}[PUe normalized IPW estimator]
Given propensity scores $e$ and PU labels, the inverse probability weighting estimator of $\Rhat_{PUe}(g)$ is
\begin{equation}
\Rhat_{PUe}(g)
=\pi\Rhat_P^{e}(g,+1)+\Rhat_U(g,-1)-\pi\Rhat_P^{e}(g,-1).
\label{eq:pue-risk}
\end{equation}
\end{definition}

\begin{figure}[t]
\centering
\includegraphics[width=0.82\linewidth]{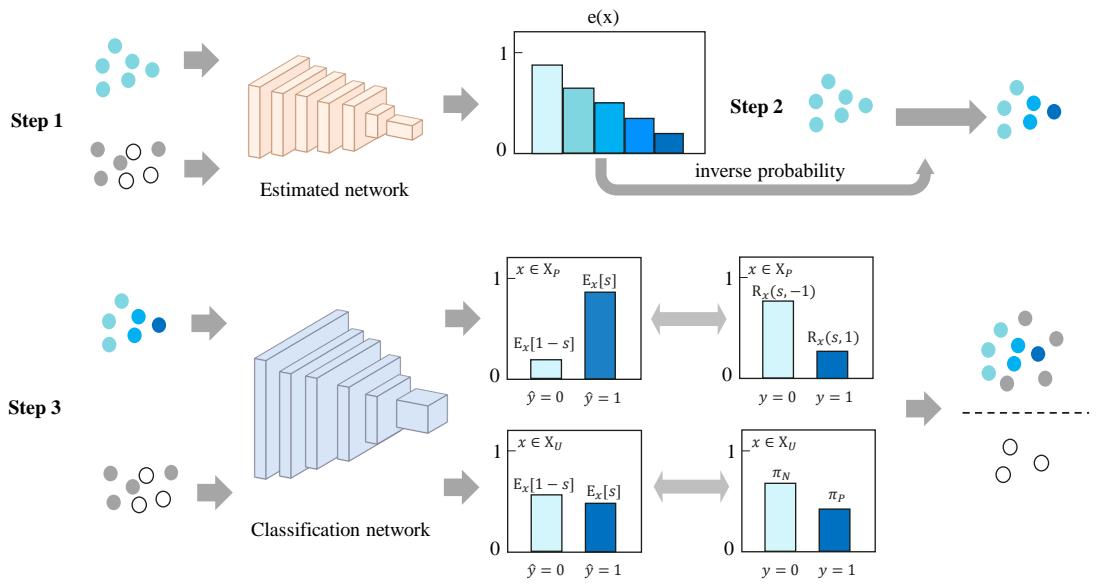}
\caption{Overview of the PUe framework. The goal is to estimate the propensity score of labeled samples and modify sample weights using normalized inverse probability weighting to obtain a loss-function estimator for biased PU learning.}
\label{fig:pue-framework}
\end{figure}

The corresponding unnormalized known-propensity IPW estimator is unbiased under the SAR-PU analysis of Bekker et al.~\cite{bekker2019beyond}. PUe uses normalized inverse-propensity weights rather than unnormalized reciprocal weights. Therefore, for the normalized PUe estimator, we state the effect of normalization as a controlled deviation. Let $h_g(x)=L(g(x),+1)-L(g(x),-1)$, $r_i=1/e(x_i)$, and $W_e=\sum_{i=1}^{n}s_ir_i$. The normalized positive correction and the oracle-normalized IPW correction satisfy
\[
\mu_e(h_g)=\frac{\sum_{i=1}^{n}s_ir_i h_g(x_i)}{W_e},
\qquad
\mu_N(h_g)=\frac{1}{N_P}\sum_{i=1}^{n}s_ir_i h_g(x_i),
\]
and therefore
\[
\mu_e(h_g)-\mu_N(h_g)
=\left(\frac{N_P-W_e}{W_e}\right)\mu_N(h_g).
\]
Thus, when $|L(g(x),y)|\le L_{\max}$ and $W_e>0$,
\[
\left|\mu_e(h_g)-\mu_N(h_g)\right|
\le 2L_{\max}\frac{|W_e-N_P|}{W_e}.
\]
This decomposition shows that the additional effect introduced by normalization is controlled by the deviation between the sample normalizer $W_e$ and the positive-population normalizer $N_P$. When this normalizer gap vanishes, the normalization-induced deviation also vanishes. Appendix~\ref{subsec:normalized-weight-decomposition} extends this calculation to estimated propensity scores.

\begin{theorem}[Known-propensity IPW error bound, \cite{bekker2019beyond}\protect\footnote{This theorem was originally published by \cite{bekker2019beyond} as Proposition 1 (Propensity-Weighted Estimator Bound), but was adapted for the notation used in this paper.}]
Let $\Rhat_{IPW}(g)$ denote the corresponding unnormalized known-propensity IPW estimator. For any predicted classes $\widehat{y}$ and real labels $y$, with probability at least $1-\eta$, $\Rhat_{IPW}(g)$ does not differ from the true PN loss $R_{PN}(g\mid y)$ by more than
\begin{equation}
\left|R_{PN}(g\mid y)-\Rhat_{IPW}(g)\right|
\le
\sqrt{\frac{L_{\max}^{2}\ln(2/\eta)}{2n}}.
\label{eq:pue-bound}
\end{equation}
Here $L_{\max}$ is the maximum absolute value of the cost function $L(g,y)$.
\end{theorem}

Using the Bekker et al.~\cite{bekker2019beyond}-adapted known-propensity IPW bound above as a reference point, the next result compares the IPW correction used by PUe with the ordinary PU estimator under biased labeling.

\begin{theorem}[Error bounds of common PU algorithm in biased scenarios]
For any predicted classes $\widehat{y}$ and real labels $y$, with probability at least $1-\eta$, the common PU estimator $\Rhat_{PU}(g)$ in biased scenarios satisfies
\begin{equation}
\left|R_{PN}(g\mid y)-\Rhat_{PU}(g)\right|
\le
2\pi L_{\max}\frac{n_U}{n_U+n_P}
+\sqrt{\frac{L_{\max}^{2}\ln(2/\eta)}{2n}}.
\label{eq:common-pu-bound}
\end{equation}
\end{theorem}
In biased scenarios, this bound is larger than the known-propensity IPW reference bound, apart from the controlled normalization term above.

\paragraph{Case 2: estimated propensity scores.}
In practice, the probability of a sample being labeled is usually unknown. We therefore estimate the propensity score:
\begin{equation}
\Rhat_{PU\widehat{e}}(g)
=\pi\Rhat_P^{\widehat{e}}(g,+1)+\Rhat_U(g,-1)-\pi\Rhat_P^{\widehat{e}}(g,-1).
\label{eq:pue-est-risk}
\end{equation}
The following bias expression is adapted from the estimated-propensity-score analysis of Bekker et al.~\cite{bekker2019beyond}, while here we study it under normalized weights and deep propensity-score estimation. Let $\omega_i$ and $\widehat{\omega}_i$ denote the corresponding normalized inverse weights computed from $e(x_i)$ and $\widehat{e}(x_i)$:
\begin{equation}
\operatorname{bias}\left(\Rhat_{PU\widehat{e}}(g)\right)
=
\pi\sum_{i=1}^{n}y_i
\left[
\left(
\frac{1}{N_P}
-\frac{\widehat{\omega}_i}{N_P\omega_i}
\right)
\left(L(g(x_i),+1)-L(g(x_i),-1)\right)
\right].
\label{eq:estimated-bias}
\end{equation}
From the bias, propensity scores need to be accurate for positive examples. When an incorrect propensity score is close to 0 or 1, especially close to 0, the bias can become large. Underestimated propensity scores are expected to result in a model with higher bias.

We estimate propensity scores by training a binary network and adding regularization:
\begin{equation}
\widehat{e}(x)=
\arg\min_{e}
\frac{\pi_1}{n_P}\sum_{i=1}^{n_P}L(e(x_i^P),+1)
+
\frac{1-\pi_1}{n_U}\sum_{i=1}^{n_U}L(e(x_i^U),-1)
+
\alpha_e\left|\sum_{x_i\in\chi_P\cup\chi_U}e(x_i)-n_P\right|,
\label{eq:propensity-estimation}
\end{equation}
where $\pi_1=n_P/(n_P+n_U)$. The regularization prevents degenerate estimates in which labeled samples are close to 1 and unlabeled samples are close to 0.

According to \cite{gerych2022recovering,bekker2019beyond}, one can use propensity scores for classification through
\[
P(\widehat{Y}\mid E,L)
=\frac{1}{n}\sum_{i=1}^{n}
s_i\left(\frac{1}{e_i}\delta_1+\left(1-\frac{1}{e_i}\right)\delta_0\right)
+(1-s_i)\delta_0.
\]
However, underestimated propensity scores reduce classification performance. PUe combines normalized inverse-propensity weights with cost-sensitive PU learning. The basic PUe loss is
\begin{equation}
\Rhat_{PU\widehat{e}}(g)
=\pi\Rhat_P^{\widehat{e}}(g,+1)+\Rhat_U(g,-1)-\pi\Rhat_P^{\widehat{e}}(g,-1).
\label{eq:pue-basic-loss}
\end{equation}
The algorithm first estimates positive-sample propensity scores, then modifies positive-sample weights with normalized propensity scores, and finally applies a PU algorithm with the modified weights. We also consider selectively labeled negative classes through PUbNe.

\section{Experiment}

We experiment on several common PU methods and compare PUe against the corresponding baselines.

\subsection{Experimental settings}

\textbf{Datasets.} We use MNIST for parity classification, CIFAR-10~\cite{krizhevsky2009learning} for vehicle class recognition, and the Alzheimer's dataset from Kaggle for real-world evaluation. On simulated MNIST and CIFAR-10 settings, the true propensity score is known a priori, allowing comparison with oracle-propensity variants.

\textbf{Baselines.} We mainly consider uPU~\cite{duplessis2014analysis}, nnPU~\cite{kiryo2017positive}, PUbN~\cite{hsieh2019classification}, and Dist-PU~\cite{zhao2022distpu}.

\textbf{Evaluation metrics.} We report accuracy (ACC), precision (Prec.), recall (Rec.), F1, area under the ROC curve (AUC), and average precision (AP). Experiments are repeated with six random seeds; means and standard deviations are reported.

\textbf{Implementation details.} All experiments are run in PyTorch. The batch size is 256 for MNIST and CIFAR-10, and 128 for Alzheimer. We use Adam with cosine annealing, initial learning rate $5\times 10^{-3}$, and weight decay $5\times 10^{-3}$. PU methods first use a 60-epoch warm-up phase and then train another 60 epochs; $\alpha$ is searched in $[0,20]$.

\begin{table}[t]
\centering
\caption{Comparative results on MNIST, CIFAR-10, and Alzheimer.}
\label{tab:main-results}
\scriptsize
\resizebox{\textwidth}{!}{
\begin{tabular}{llrrrrrr}
\toprule
Dataset & Method & ACC (\%) & Prec. (\%) & Rec. (\%) & F1 (\%) & AUC (\%) & AP (\%) \\
\midrule
\multirow{12}{*}{MNIST} & uPU & 85.17 (1.90) & 78.38 (2.36) & 96.69 (1.33) & 86.55 (1.50) & 85.34 (1.87) & 77.41 (2.26)\\
& nnPU & 87.81 (1.65) & 83.78 (2.14) & 93.40 (1.68) & 88.31 (1.51) & 87.89 (1.63) & 81.62 (2.28)\\
& PUbN & 87.07 (1.73) & 83.43 (2.74) & 92.21 (2.55) & 87.55 (1.55) & 87.14 (2.74) & 80.70 (2.59)\\
& Dist-PU & 89.79 (0.59) & 92.22 (1.67) & 88.86 (1.48) & 90.49 (0.47) & 96.60 (0.30) & 96.92 (0.21)\\
& uPUe & 92.28 (1.32) & 89.41 (1.87) & 95.71 (1.43) & 92.44 (1.25) & 92.33 (1.31) & 87.69 (1.92)\\
& nnPUe & 92.45 (1.61) & 90.45 (2.26) & 94.73 (1.24) & 92.53 (1.55) & 92.48 (1.60) & 88.29 (2.43)\\
& PUbNe & 94.70 (0.63) & 97.16 (1.59) & 91.33 (1.58) & 94.46 (0.71) & 94.66 (0.50) & 93.18 (0.55)\\
& Dist-PUe & 92.57 (0.78) & 93.02 (0.86) & 90.58 (1.47) & 92.77 (0.84) & 97.13 (0.57) & 97.43 (0.44)\\
& uPUe' & 92.39 (1.36) & 89.33 (1.87) & 96.07 (1.70) & 92.56 (1.30) & 92.44 (1.87) & 87.76 (1.93)\\
& nnPUe' & 92.94 (1.27) & 91.11 (1.84) & 94.98 (2.38) & 92.98 (1.29) & 92.96 (1.27) & 89.00 (1.77)\\
& PUbNe' & 94.48 (0.93) & 97.74 (0.46) & 90.90 (2.10) & 94.18 (1.06) & 94.43 (0.95) & 93.45 (1.00)\\
& Dist-PUe' & 91.73 (0.95) & 93.34 (0.42) & 89.60 (2.38) & 91.41 (1.11) & 96.48 (1.54) & 96.60 (1.37)\\
\midrule
\multirow{12}{*}{CIFAR-10} & uPU & 76.33 (1.76) & 83.59 (0.65) & 50.81 (5.57) & 63.01 (4.35) & 84.40 (3.12) & 78.68 (2.22)\\
& nnPU & 82.47 (0.69) & 75.30 (1.35) & 83.68 (1.64) & 79.25 (0.75) & 90.01 (0.73) & 85.22 (0.73)\\
& PUbN & 84.54 (0.54) & 82.18 (1.38) & 78.40 (2.70) & 80.20 (0.99) & 91.62 (0.27) & 87.31 (0.40)\\
& Dist-PU & 84.45 (1.00) & 80.90 (1.51) & 80.04 (1.62) & 80.46 (1.26) & 90.89 (1.20) & 85.67 (1.54)\\
& uPUe & 79.31 (0.90) & 81.37 (2.16) & 62.86 (5.02) & 70.73 (2.58) & 86.76 (0.82) & 80.07 (1.56)\\
& nnPUe & 83.33 (1.04) & 76.83 (2.75) & 83.88 (2.86) & 80.11 (0.85) & 90.53 (0.79) & 85.33 (1.36)\\
& PUbNe & 85.66 (0.58) & 83.18 (1.63) & 80.52 (2.29) & 81.79 (0.66) & 92.69 (0.74) & 89.41 (1.24)\\
& Dist-PUe & 86.97 (0.62) & 82.11 (1.69) & 86.46 (2.00) & 84.19 (0.76) & 93.50 (0.66) & 89.38 (1.78)\\
& uPUe' & 79.11 (1.75) & 85.11 (1.58) & 57.92 (4.96) & 68.80 (3.61) & 86.57 (1.56) & 81.56 (2.18)\\
& nnPUe' & 84.24 (0.67) & 78.40 (0.86) & 83.66 (2.35) & 80.92 (1.02) & 91.35 (0.99) & 87.15 (1.83)\\
& PUbNe' & 87.02 (0.47) & 83.84 (1.44) & 83.78 (2.68) & 83.76 (0.82) & 93.80 (0.42) & 90.69 (0.65)\\
& Dist-PUe' & 86.17 (2.52) & 80.97 (3.20) & 85.57 (3.32) & 83.19 (3.06) & 92.65 (2.20) & 88.22 (3.19)\\
\midrule
\multirow{8}{*}{Alzheimer} & uPU & 63.46 (1.72) & 66.80 (3.45) & 55.06 (9.83) & 59.65 (4.70) & 68.08 (2.16) & 67.78 (2.22)\\
& nnPU & 69.19 (0.85) & 65.46 (1.06) & 81.46 (5.22) & 72.48 (1.62) & 73.50 (0.84) & 68.74 (1.83)\\
& PUbN & 68.75 (0.96) & 66.65 (2.04) & 75.64 (5.58) & 70.68 (1.52) & 72.57 (0.63) & 66.99 (0.80)\\
& Dist-PU & 69.74 (0.50) & 67.74 (1.57) & 75.69 (4.72) & 71.37 (1.33) & 74.34 (1.10) & 69.53 (1.39)\\
& uPUe & 65.30 (3.36) & 68.53 (1.99) & 56.73 (11.93) & 61.32 (7.58) & 70.49 (3.99) & 70.28 (3.09)\\
& nnPUe & 69.38 (0.56) & 65.97 (0.68) & 80.06 (2.77) & 72.30 (0.91) & 74.49 (0.34) & 69.98 (0.60)\\
& PUbNe & 69.52 (0.85) & 68.28 (1.28) & 72.95 (1.99) & 70.51 (0.82) & 74.68 (1.08) & 70.33 (0.98)\\
& Dist-PUe & 70.51 (0.62) & 68.91 (1.00) & 75.02 (2.26) & 71.81 (0.83) & 75.19 (0.69) & 71.23 (2.08)\\
\bottomrule
\end{tabular}}
\end{table}

\subsection{Comparison with state-of-the-art methods}

The results on all datasets are recorded in Table~\ref{tab:main-results}. In most metrics, the proposed PUe variants outperform the corresponding competitors on biased datasets, improving the original PU method by about 1\% to 5\%. A model using known propensity scores is not necessarily the best; a model with estimated propensity scores can perform better in many cases, which is consistent with observations in causal inference~\cite{hirano2003efficient}. In the ablation study, the performance of PUe remains comparable to that of the advanced PU algorithm even when labels are evenly distributed.

\subsection{Ablation studies}

\begin{figure}[t]
\centering
\includegraphics[width=0.72\linewidth]{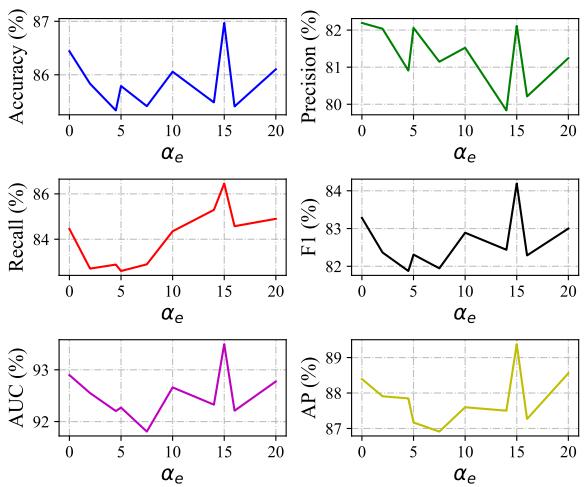}
\caption{Influence of different $\alpha_e$ values on CIFAR-10 with Dist-PUe.}
\label{fig:alpha}
\end{figure}

\textbf{Effectiveness of hyper-parameters.} Ablation experiments verify the validity of hyperparameters, as shown in Figure~\ref{fig:alpha}. PUe is sensitive to $\alpha_e$ and does not change monotonically. In the original experiment, $\alpha_e=15$ gives the best performance.

\textbf{Effectiveness of labeled-sample distribution.} Table~\ref{tab:ablation} shows that label-distribution deviation significantly affects the improvement from PUe. In general, when labeled samples are more biased, PUe provides larger improvements. When deviation is extremely large, the improvement weakens because some classes have too few labeled samples, making propensity scores close to 0 and affecting model performance.

\begin{table}[t]
\centering
\caption{Ablation results on CIFAR-10 with checkmark indicating the enabling of the corresponding regularization loss term and different labeled distributions.}
\label{tab:ablation}
\scriptsize
\resizebox{\textwidth}{!}{
\begin{tabular}{llrrrrrrr}
\toprule
Method & $\alpha_e$ & Labeled distribution & ACC (\%) & Prec. (\%) & Rec. (\%) & F1 (\%) & AUC (\%) & AP (\%)\\
\midrule
Dist-PUe & \checkmark & [.25,.25,.25,.25] & 91.79 (0.86) & 89.02 (2.63) & 90.78 (1.81) & 89.84 (0.92) & 97.11 (0.52) & 95.38 (1.18)\\
Dist-PUe &  & [.25,.25,.25,.25] & 91.30 (0.80) & 89.22 (1.56) & 89.05 (1.93) & 89.11 (1.03) & 96.64 (0.53) & 94.85 (0.77)\\
Dist-PU &  & [.25,.25,.25,.25] & 91.88 (0.52) & 89.87 (1.09) & 89.84 (0.81) & 89.85 (0.62) & 96.92 (0.45) & 95.49 (0.72)\\
Dist-PUe & \checkmark & [.10,.10,.30,.50] & 91.50 (0.32) & 89.07 (1.15) & 89.78 (0.97) & 89.41 (0.34) & 96.67 (0.14) & 95.64 (0.26)\\
Dist-PUe &  & [.10,.10,.30,.50] & 90.40 (0.66) & 88.86 (3.15) & 87.14 (0.74) & 87.89 (0.74) & 95.71 (0.40) & 94.19 (0.65)\\
Dist-PU &  & [.10,.10,.30,.50] & 90.79 (0.72) & 91.31 (1.54) & 85.12 (2.44) & 88.07 (1.07) & 95.81 (0.83) & 91.28 (2.35)\\
Dist-PUe & \checkmark & [.72,.15,.10,.03] & 86.97 (0.62) & 82.11 (1.69) & 86.46 (2.00) & 84.19 (0.76) & 93.50 (0.66) & 89.38 (1.78)\\
Dist-PUe &  & [.72,.15,.10,.03] & 86.44 (0.67) & 82.19 (1.54) & 84.45 (1.78) & 83.28 (0.82) & 92.90 (0.63) & 88.39 (1.32)\\
Dist-PU &  & [.72,.15,.10,.03] & 84.45 (1.00) & 80.90 (1.51) & 80.04 (1.62) & 80.46 (1.26) & 90.89 (1.20) & 85.67 (1.54)\\
Dist-PUe & \checkmark & [.08,.86,.02,.04] & 88.15 (1.22) & 88.05 (2.69) & 81.70 (5.41) & 84.56 (2.23) & 94.46 (1.01) & 92.79 (1.09)\\
Dist-PUe &  & [.08,.86,.02,.04] & 86.39 (2.11) & 84.47 (3.51) & 81.05 (4.32) & 82.62 (2.77) & 92.42 (2.25) & 89.84 (2.40)\\
Dist-PU &  & [.08,.86,.02,.04] & 86.74 (1.58) & 86.56 (2.51) & 79.30 (4.60) & 82.65 (2.48) & 93.30 (1.32) & 91.08 (1.69)\\
\bottomrule
\end{tabular}}
\end{table}

\section{Conclusion}

This paper proposes PUe, a PU learning method from the perspective of propensity scores for PU learning with biased labels in deep learning. PUe enhances original cost-sensitive PU algorithms, improves prediction precision under biased sample labeling, and has the degenerate ability that prediction precision under unbiased labeling is not lower than that of the original PU algorithm. PUe consistently outperforms state-of-the-art methods on most metrics on biased labeled datasets, including MNIST, CIFAR-10, and Alzheimer's. We hope that the proposed propensity-score estimation scheme for deep learning can also provide inspiration for other weakly supervised scenarios, especially those where label distribution is unknown.

\section*{Acknowledgement}

This work was supported by Huawei Noah's Ark Lab.

\bibliographystyle{unsrt}
\bibliography{references}

\appendix

\section{Algorithm}

\begin{table}[htbp]
\centering
\caption{PUe algorithm.}
\label{alg:pue}
\begin{tabular}{p{0.94\linewidth}}
\toprule
\textbf{Require:} data $\chi_P,\chi_U$, sizes $n,n_P,n_U$, hyperparameters $\alpha_e,\pi$.\\
1. Estimate $\widehat{e}(x)$ by minimizing Eq.~\eqref{eq:propensity-estimation}.\\
2. Compute the weight of labeled samples: $w_i^P=\pi\widehat{\omega}_i^P$.\\
3. For each training iteration, shuffle $(\chi_P,\chi_U)$ into mini-batches.\\
4. For each mini-batch $(\chi_j^P,\chi_j^U)$, compute the corresponding $\Rhat_{PUe}(g)$.\\
5. Update $\theta$ with gradient information $\nabla_{\theta}\Rhat_{PUe}(g)$.\\
\bottomrule
\end{tabular}
\end{table}

\section{Experiment details}

\begin{table}[htbp]
\centering
\caption{Summary of used datasets and their corresponding models.}
\label{tab:dataset-summary}
\scriptsize
\resizebox{\textwidth}{!}{
\begin{tabular}{llrrrrlll}
\toprule
Dataset & Input size & $n_P$ & $n_U$ & Testing & $\pi_P$ & Positive class & True $e(x)$ & Model\\
\midrule
MNIST & $28\times 28$ & 2500 & 60000 & 10000 & 0.5 & Even (0, 2, 4, 6, 8) & [.65,.15,.1,.07,.03] & 6-layer MLP\\
CIFAR-10 & $3\times 32\times 32$ & 1000 & 50000 & 10000 & 0.4 & Vehicles (0, 1, 8, 9) & [.72,.15,.1,.03] & 13-layer CNN\\
Alzheimer & $3\times 224\times 224$ & 769 & 5121 & 1279 & 0.5 & Alzheimer's disease & unknown & ResNet-50\\
\bottomrule
\end{tabular}}
\end{table}

\section{Complementary experiment}

\begin{table}[htbp]
\centering
\caption{Supplemental experiments on MNIST.}
\label{tab:supp-mnist}
\scriptsize
\resizebox{\textwidth}{!}{
\begin{tabular}{llrrrrrr}
\toprule
Method & Labeled distribution & ACC (\%) & Prec. (\%) & Rec. (\%) & F1 (\%) & AUC (\%) & AP (\%)\\
\midrule
LRe & [.65,.15,.10,.07,.03] & 86.19 (0.75) & 92.94 (0.64) & 77.89 (1.38) & 84.75 (0.93) & 88.06 (0.93) & 88.72 (1.09)\\
nnPUe & [.65,.15,.10,.07,.03] & 92.45 (1.61) & 90.45 (2.26) & 94.73 (1.24) & 92.53 (1.55) & 92.48 (1.60) & 88.29 (2.43)\\
nnPU without normalize & [.65,.15,.10,.07,.03] & 90.95 (1.61) & 88.18 (2.40) & 94.38 (2.74) & 91.13 (1.56) & 91.00 (1.61) & 85.98 (2.25)\\
Self-PU & [.65,.15,.10,.07,.03] & 90.08 (0.47) & 90.08 (0.47) & 89.35 (1.17) & 90.70 (1.73) & 90.00 (0.53) & 85.61 (0.69)\\
anchor & [.65,.15,.10,.07,.03] & 88.22 (0.95) & 94.66 (1.41) & 80.70 (3.15) & 87.06 (1.37) & 92.36 (1.92) & 93.37 (2.33)\\
\bottomrule
\end{tabular}}
\end{table}

LRe denotes logistic-regression estimation of propensity scores for PU learning. According to Gerych et al.~\cite{gerych2022recovering}, identifiable propensity-score estimation requires assumptions about the data. The results above show that PUe improves over self-supervised and linear-estimation baselines in the biased-label setting.

\section{Additional Discussion of Related SAR-PU Work}

This appendix summarizes the relationship between PUe and the SAR-PU framework of Bekker et al.~\cite{bekker2019beyond}. The main text identifies Bekker et al.~\cite{bekker2019beyond} as the closest prior work; Table~\ref{tab:sar-pu-comparison} provides a compact component-level comparison.

\begin{table}[htbp]
\centering
\caption{Component-level comparison with related SAR-PU work.}
\label{tab:sar-pu-comparison}
\scriptsize
\begin{tabularx}{\textwidth}{>{\raggedright\arraybackslash}p{0.24\textwidth}YY}
\toprule
Aspect & Bekker et al. (2019) & PUe\\
\midrule
Problem setting & SAR positive-unlabeled learning with feature-dependent labeling & Builds on the SAR-PU setting for deep biased PU learning\\
Propensity score & Defines and uses propensity scores for biased PU learning & Uses the same foundation with normalized weights in PUe notation\\
Estimator & Propensity-weighted SAR-PU estimator & Reformulates and normalizes the estimator in PN/PU risk notation\\
Known-propensity IPW identity and bound & Analyzes unbiasedness with known propensity scores and gives a deviation bound & Uses this analysis as the IPW reference and adds normalized-weight deviation analysis\\
Estimated-score bias & Analyzes bias caused by estimated propensity scores & Adapts the bias expression to normalized weights and deep estimation\\
Additional theory & Establishes foundational SAR-PU estimator analysis & Analyzes normalized sample-weight error and common PU estimators under biased labeling\\
Propensity estimation & Studies estimation within the SAR-PU framework & Uses deep neural estimation with an added regularization term\\
Cost-sensitive integration & Applies propensity weighting in the SAR-PU setting & Couples normalized weighting with uPU, nnPU, PUbN, and Dist-PU\\
Labeling scope & Models selection of positive labels & Extends the framework to selectively labeled negatives through PUbNe\\
Empirical scope & Classical SAR-PU evaluation & Evaluates on MNIST, CIFAR-10, and ADNI\\
\bottomrule
\end{tabularx}
\end{table}

\section{Proofs}
\label{sec:proofs}

\subsection{Error bound of bias}

Assume that the propensity-score estimate has a maximum error ratio $\beta$, with
\[
\beta e(x_i^L)\le \widehat{e}(x_i^L)\le e(x_i^L).
\]
The sample $x_i^L$ has weight $1/(n\widehat{e}(x_i^L))$, whose error is bounded by
\[
\operatorname{bias}\left(\frac{1}{n\widehat{e}(x_i^L)}\right)
\le
\frac{1}{ne(x_i^L)}\left(\frac{1}{\beta}-1\right).
\]

\subsection{Error ratio}

In PUe, sample $x_i^L$ has normalized weight
\[
\pi\frac{1/\widehat{e}(x_i^L)}{\sum_j1/\widehat{e}(x_j^L)}.
\]
Let $P(\gamma e(x_i^L)<\widehat{e}(x_i^L)\le e(x_i^L))=\alpha$ for $S_1$ and $P(\beta e(x_i^L)<\widehat{e}(x_i^L)\le \gamma e(x_i^L))=1-\alpha$ for $S_2$, where $\beta<\gamma<1$. If
\[
\sum_{i\in S_1}\frac{1}{e(x_i^L)}
=
\sum_{i\in S_2}\frac{1}{e(x_i^L)}
=B,
\]
then $\sum_j1/e(x_j^L)=2B=N_P$. For $x_i^L\in S_1$,
\[
\frac{1}{e(x_i^L)}\le \frac{1}{\widehat{e}(x_i^L)}<\frac{1}{\gamma e(x_i^L)},
\]
and for $x_i^L\in S_2$,
\[
\frac{1}{\gamma e(x_i^L)}\le \frac{1}{\widehat{e}(x_i^L)}<\frac{1}{\beta e(x_i^L)}.
\]
Thus
\[
B\left(1+\frac{1}{\gamma}\right)
\le
\sum_j\frac{1}{\widehat{e}(x_j^L)}
<
B\left(\frac{1}{\gamma}+\frac{1}{\beta}\right),
\]
which bounds the normalized sample-weight error and shows that normalization reduces the error ratio relative to unnormalized reciprocal weights.

\subsection{Expectation}

According to the propensity-score definition, each labeled positive sample $x_j^P$ corresponds in expectation to $1/e(x_j^P)$ positive samples. Because $P(x\mid s=1)=P(x,y=1\mid s=1)$,
\[
\begin{aligned}
\mathbb{E}_{P(x\mid s=1)}
\frac{1}{P(s=1\mid x,y=1)}
&=
\sum_x P(x,y=1\mid s=1)\frac{1}{P(s=1\mid x,y=1)}\\
&=
\sum_x \frac{P(s=1\mid x,y=1)P(x,y=1)}{P(s=1)}
\frac{1}{P(s=1\mid x,y=1)}\\
&=
\sum_x \frac{P(x,y=1)}{P(s=1)}
=\frac{n}{n_P}\sum_xP(x,y=1)
=\frac{N_P}{n_P}.
\end{aligned}
\]
This indicates that $\sum_{j=1}^{n_P}1/e(x_j^P)=N_P$.

\subsection{PUbN}

Let $\sigma(x)=p(s=+1\mid x)$, which is unknown and replaced by $\widehat{\sigma}(x)$. The PUbN risk is
\[
R_{PUbN}(g)=\pi R_P(g,+1)+\rho R_{bN}(g,-1)+\overline{R}_{s=-1,\eta,\widehat{\sigma}}(g),
\]
where
\[
\begin{aligned}
\overline{R}_{s=-1,\eta,\widehat{\sigma}}(g)
&=
\mathbb{E}_{x\sim p(x)}
\left[\mathbb{1}_{\widehat{\sigma}(x)\le\eta}L(-g(x))(1-\widehat{\sigma}(x))\right]\\
&\quad+\pi\mathbb{E}_{x\sim p_P(x)}
\left[\mathbb{1}_{\widehat{\sigma}(x)>\eta}L(-g(x))
\frac{1-\widehat{\sigma}(x)}{\widehat{\sigma}(x)}\right]\\
&\quad+\rho\mathbb{E}_{x\sim p_{bN}(x)}
\left[\mathbb{1}_{\widehat{\sigma}(x)>\eta}L(-g(x))
\frac{1-\widehat{\sigma}(x)}{\widehat{\sigma}(x)}\right].
\end{aligned}
\]
The corresponding empirical risk is
\[
\Rhat_{PUbN,\eta,\widehat{\sigma}}(g)
=
\pi\Rhat_P(g,+1)+\rho\Rhat_{bN}(g,-1)
+\widehat{\overline{R}}_{s=-1,\eta,\widehat{\sigma}}(g).
\]

\subsection{PUbNe}

The PUbNe risk is
\[
\Rhat_{PUbN\widehat{e},\eta,\widehat{\sigma}}(g)
=
\pi\Rhat_P^{\widehat{e}}(g,+1)
+\rho\Rhat_{bN}^{\widehat{e}}(g,-1)
+\widehat{\overline{R}}_{s=-1,\eta,\widehat{\sigma}}^{\widehat{e}}(g),
\]
where
\[
\Rhat_{bN}^{\widehat{e}}(g,-1)
=
\sum_{i=1}^{n_{bN}}
\widehat{\omega}_i^{bN}L(g(x_i^{bN}),-1).
\]

\subsection{Normalized-weight decomposition}
\label{subsec:normalized-weight-decomposition}

Let $h_g(x)=L(g(x),+1)-L(g(x),-1)$, $r_i=1/e(x_i)$, and $\widehat{r}_i=1/\widehat{e}(x_i)$. For the known-propensity normalizer $W_e=\sum_{i=1}^{n}s_ir_i$, define
\[
\mu_e(h_g)=\frac{\sum_{i=1}^{n}s_ir_i h_g(x_i)}{W_e},
\qquad
\mu_N(h_g)=\frac{1}{N_P}\sum_{i=1}^{n}s_ir_i h_g(x_i).
\]
Then
\[
\begin{aligned}
\mu_e(h_g)-\mu_N(h_g)
&=
\left(\frac{1}{W_e}-\frac{1}{N_P}\right)
\sum_{i=1}^{n}s_ir_i h_g(x_i)\\
&=
\left(\frac{N_P-W_e}{W_e}\right)\mu_N(h_g).
\end{aligned}
\]
If $|L(g(x),y)|\le L_{\max}$, then $|h_g(x)|\le 2L_{\max}$ and
\[
\left|\mu_e(h_g)-\mu_N(h_g)\right|
\le
2L_{\max}\frac{|W_e-N_P|}{W_e}.
\]

For estimated propensity scores, let $\Delta_i=\widehat{r}_i-r_i$ and $\widehat{W}=W_e+\sum_{i=1}^{n}s_i\Delta_i$. The estimated normalized correction
\[
\widehat{\mu}(h_g)
=
\frac{\sum_{i=1}^{n}s_i\widehat{r}_ih_g(x_i)}{\widehat{W}}
\]
satisfies the exact identity
\[
\widehat{\mu}(h_g)-\mu_e(h_g)
=
\frac{\sum_{i=1}^{n}s_i\Delta_i\left(h_g(x_i)-\mu_e(h_g)\right)}{\widehat{W}}.
\]
Consequently, if $\widehat{W}>0$,
\[
\left|\widehat{\mu}(h_g)-\mu_e(h_g)\right|
\le
\frac{4L_{\max}\sum_{i=1}^{n}s_i|\Delta_i|}{\widehat{W}}.
\]
This gives a PUe-specific normalized-weight error term: with normalized weights, the estimated-propensity effect depends on reciprocal-propensity errors relative to the estimated normalizer, rather than only on pointwise propensity errors.

\subsection{Known-propensity IPW identity and bounds}

The following proof is for the corresponding unnormalized known-propensity IPW estimator and follows the propensity-weighted estimator analysis of Bekker et al.~\cite{bekker2019beyond}. PUe then uses normalized weights, whose additional effect is controlled in Appendix~\ref{subsec:normalized-weight-decomposition}:
\[
\begin{aligned}
\mathbb{E}[\Rhat_{IPW}(g)]
&=
\mathbb{E}\left[
\pi\Rhat_{P,IPW}^{e}(g,+1)+\Rhat_U(g,-1)-\pi\Rhat_{P,IPW}^{e}(g,-1)
\right]\\
&=
\mathbb{E}\left[
\frac{1}{n}\sum_{i=1}^{n_P}
\frac{1}{e(x_i^P)}
\left(L(g(x_i^P),+1)-L(g(x_i^P),-1)\right)
+
\frac{1}{n}\sum_{i=1}^{n}L(g(x_i),-1)
\right]\\
&=
\frac{1}{n}\sum_{i=1}^{n}
\Bigg[
y_i e_i\frac{1}{e(x_i)}L(g(x_i),+1)\\
&\qquad
+y_ie_i\left(1-\frac{1}{e(x_i)}\right)L(g(x_i),-1)\\
&\qquad
+(1-y_ie_i)L(g(x_i),-1)
\Bigg]\\
&=
\frac{1}{n}\sum_{i=1}^{n}
\left[
y_iL(g(x_i),+1)+(1-y_i)L(g(x_i),-1)
\right]\\
&=R_{PN}(g\mid y).
\end{aligned}
\]
The change of $\Rhat_{IPW}(g)$ is no more than $L_{\max}/n$ if some $x_i\in\chi_P\cup\chi_U$ is replaced. McDiarmid's inequality gives
\[
\Pr\left\{|\Rhat_{IPW}(g)-R_{PN}(g\mid y)|\ge\epsilon\right\}
\le
2\exp\left(-\frac{2\epsilon^2}{n(L_{\max}/n)^2}\right).
\]
Setting the right side equal to $\eta$ yields
\[
\epsilon=
\sqrt{\frac{L_{\max}^{2}\ln(2/\eta)}{2n}},
\]
and hence, with probability at least $1-\eta$,
\[
\left|\Rhat_{IPW}(g)-R_{PN}(g\mid y)\right|
\le
\sqrt{\frac{L_{\max}^{2}\ln(2/\eta)}{2n}}.
\]

Because
\[
\Rhat_{IPW}(g)=\pi\Rhat_{P,IPW}^{e}(g,+1)+\Rhat_U(g,-1)-\pi\Rhat_{P,IPW}^{e}(g,-1)
\]
and
\[
\Rhat_{PU}(g)=\pi\Rhat_P(g,+1)+\Rhat_U(g,-1)-\pi\Rhat_P(g,-1),
\]
we obtain the biased-labeling comparison term
\[
|\Rhat_{IPW}(g)-\Rhat_{PU}(g)|
\le
2\pi L_{\max}\frac{n_U}{n_U+n_P}.
\]
Therefore,
\[
\left|R_{PN}(g\mid y)-\Rhat_{PU}(g)\right|
\le
2\pi L_{\max}\frac{n_U}{n_U+n_P}
+
\sqrt{\frac{L_{\max}^{2}\ln(2/\eta)}{2n}}.
\]

\end{document}